\documentclass[letterpaper, 10 pt, conference]{ieeeconf}  

\IEEEoverridecommandlockouts                              
\usepackage{amsmath} 
\usepackage{lettrine}
\usepackage{graphicx}
\usepackage{booktabs}

\title{\LARGE \bf
200x Low-dose PET Reconstruction using Deep Learning
}

\author{Junshen Xu$^\dagger $, Enhao Gong$^\dagger $, John Pauly and Greg Zaharchuk$^{*}$
\thanks{This work was supported in part by ---. Dagger indicates equal contribution of the first two authors. Asterisk indicates corresponding author.}
\thanks{$\dagger$Junshen Xu is with the Department of Electrical Engineering,Stanford University, Stanford, CA, 94305, USA and the Department of Engineering Physics, Tsinghua University, Beijing, 100084 China}
\thanks{$\dagger$Enhao Gong and John Pauly is with the Department of Electrical Engineering, Stanford University, Stanford, CA, 94305, USA}
\thanks{$*$Greg Zaharchuk is with the Department of Radiology, Stanford University, Stanford, CA, 94305, USA}
}

\begin{document}

\maketitle
\thispagestyle{empty}
\pagestyle{empty}

\begin{abstract}

Positron emission tomography (PET) is widely used in various clinical applications, including cancer diagnosis, heart disease and neuro disorders.
The use of radioactive tracer in PET imaging raises concerns due to the risk of radiation exposure. To minimize this potential risk in PET imaging, efforts have been made to reduce the amount of radio-tracer usage. However, lowing dose results in low Signal-to-Noise-Ratio (SNR) and loss of information, both of which will heavily affect clinical diagnosis. Besides, the ill-conditioning of low-dose PET image reconstruction makes it a difficult problem for iterative reconstruction algorithms. Previous methods proposed are typically complicated and slow, yet still cannot yield satisfactory results at significantly low dose. Here, we propose a deep learning method to resolve this issue with an encoder-decoder residual deep network with concatenate skip connections. Experiments shows the proposed method can reconstruct low-dose PET image to a standard-dose quality with only two-hundredth dose. Different cost functions for training model are explored. Multi-slice input strategy is introduced to provide the network with more structural information and make it more robust to noise. Evaluation on ultra-low-dose clinical data shows that the proposed method can achieve better result than the state-of-the-art methods and reconstruct images with comparable quality using only 0.5\% of the original regular dose.

\end{abstract}



\section{Introduction}
\begin{figure}[thpb]
\centerline{\includegraphics[width=0.3\textwidth]{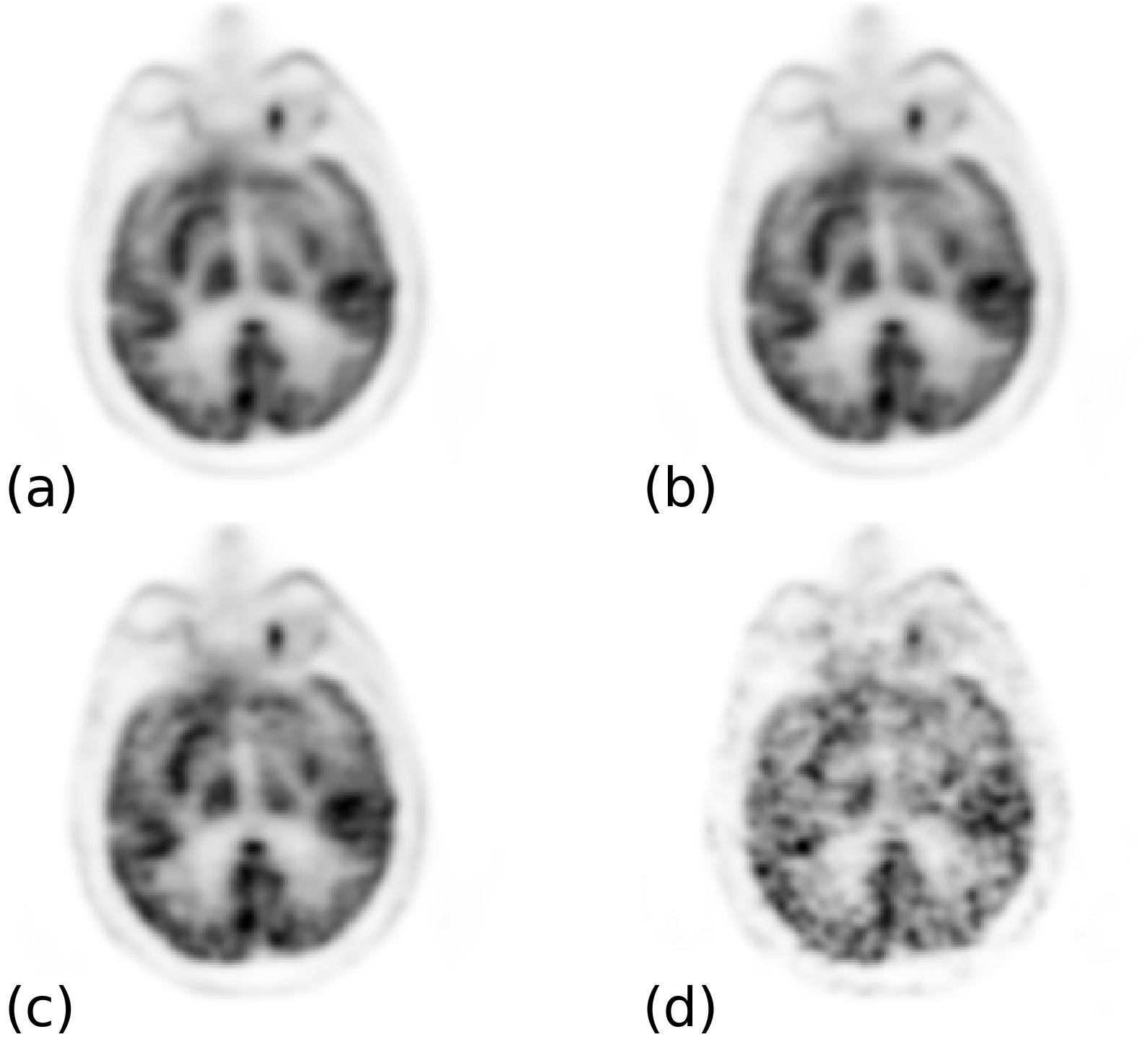}}
\caption{PET images with normal dose and different levels of dose reduction. (a) standard-dose, (b) quarter-dose, (c) twentieth-dose, and (d) two-hundredth-dose.}\label{fig_different_dose}
\label{fig}
\end{figure}

\lettrine[lines=2]{P}{ositron} emission tomography (PET) has a wide range of clinical applications, such as cancer diagnosis, tumor detection\cite{Ono2007TheScreening} and early diagnosis of neuro diseases\cite{Herholz2002DiscriminationPET}, 
for its ability of cellular level imaging and high specificity. In order to acquire high quality PET image for diagnostic purpose, a standard dose of radioactive tracer should be injected to the subject which will lead to higher risk of radiation exposure damage. Usually, a PET scan or the a scan of PET/MR and PET/CT can expose patients with even more ionizing radiation than a scan using CT along. To minimize such risk, the well-known principle of ALARA (as low as reasonably achievable)\cite{Voss2009TheOncology} is adopted in clinical practice. In addition, lowering injected dose in PET can also result in reduction of imaging costs, shorter imaging time and may improve imaging logistics when fast-decaying tracers are used. However, dose reduction will adversely affect PET image quality with lower Signal-to-Noise-Ratio (SNR), as shown in Fig. \ref{fig_different_dose}.

\begin{figure*}[thpb]
\centerline{\includegraphics[width=\textwidth]{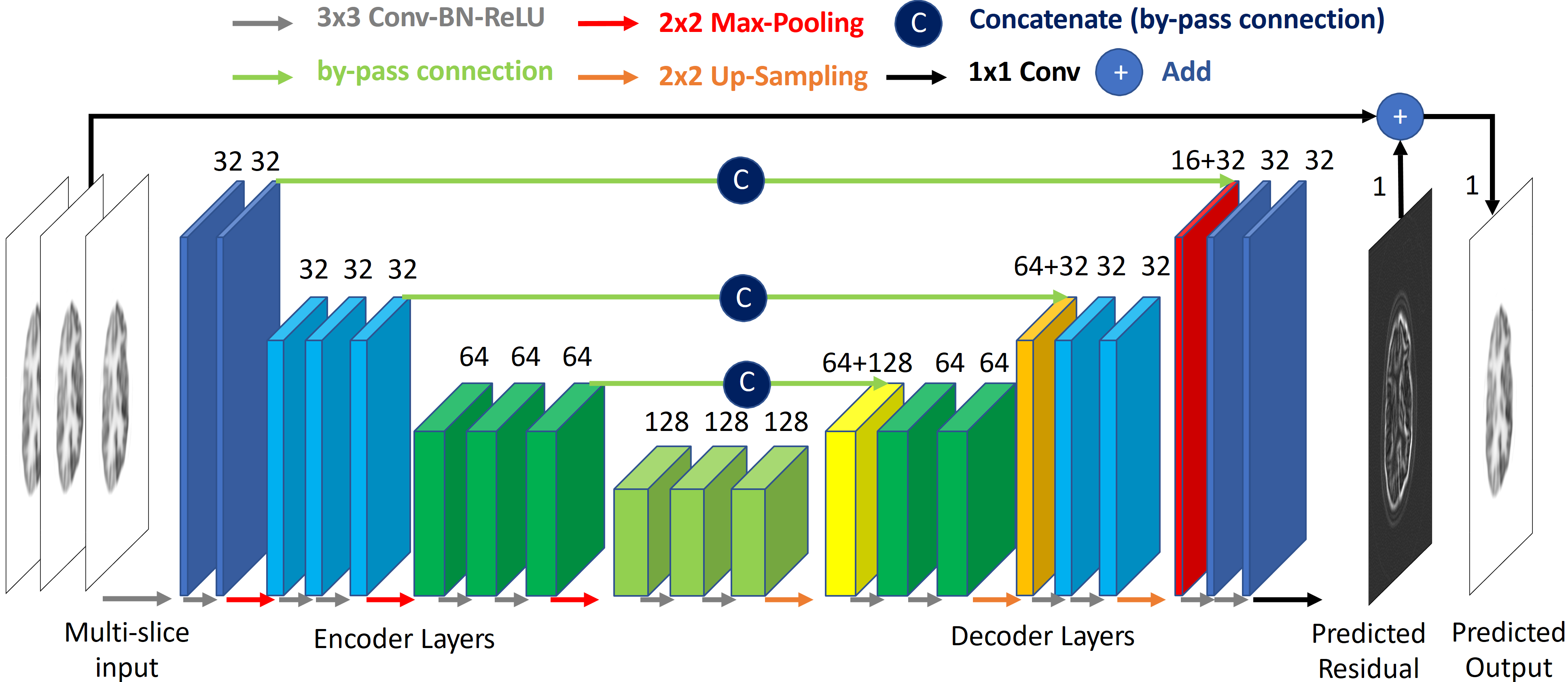}}
\caption{Overall architecture of our proposed network.}\label{fig_network_structure}
\end{figure*}

To address this problem, many algorithms were proposed to improve the image quality for low-dose PET image. In general, these algorithms can be categorized into three categories: (1) iterative reconstruction algorithm, (2) image filtering and post-processing, and (3) machine learning.

Iterative reconstruction algorithms formulate the low-dose reconstruction problem as a convex optimization problem combining statistical model of the acquired data (i.e., sinogram or listmode) and the regularization term to suppress noise. Wang \textit{et al}. \cite{ChenyeWang2014LowRegularization} proposed an iterative algorithm using a Total Variation (TV) regularization to reduce the noise of synthetic emission phantom with different photon counts. Although Iterative reconstruction algorithms are potentially most accurate since they consider the raw count information directly, they also have three main weaknesses. First, the substantial computational expenses interacting with all the acquired data make most of this kind of methods time-consuming. Second, iterative methods are typically vendor-specific, since different scanners may adopt different geometric configurations, data formats (e.g., time-of-flight (TOF) \cite{Moses2003TimeRevisited} and depth-of-interaction (DOI) \cite{Ohi2007InvestigationSystem}), and data correction procedures, which will significantly affect the raw data. Finally, in these methods, a predefined regularization term is need, which may leads to undesirable over-smoothing, artifacts or hallucinated textures.

As for image processing methods, several general-purpose image denoising algorithms, such as nonlocal means (NLM) \cite{Buades2005ADenoising} and block-matching 3D (BM3D) \cite{Dabov2006ImageFiltering}, are introduced into PET image denoising \cite{Yu2016NoiseImages,DuttaJoyitaANDLeahy2013Non-LocalImages}. Besides, Bagci \textit{et. al.} \cite{Bagci2013DenoisingEstimate} combined singular value thresholding method and Stein's unbiased risk estimate to denoise PET image. Based on the multi-scale Curvelet and Wavelet analysis, Pogam \textit{et. al.} \cite{Pogam2013DenoisingQuantitation} proposed a method to denoise PET image while preserving image resolution and quantification. 

Another important category is the data-driven machine learning methods such as mapping-based sparse representation \cite{Wang2016PredictingRepresentation}, semi-supervised tripled dictionary \cite{Wang2017SemisupervisedMRI}, and multi-level canonical correlation analysis \cite{An2016Multi-LevelEstimation}. Instead of denoise the low-dose PET image directly, machine learning methods utilize paired low-dose and standard-dose images to train models that can predicts standard-dose images from low-dose inputs.

Recently, deep learning attracts a lot of attention in computer vision applications, yields much better results compared with traditional methods, and achieves human-level performance in some tasks such as image classification \cite{He2015DelvingClassification} and face verification \cite{Taigman2014DeepFace:Verification}. Several key factors contribute to the success of deep learning methods:  (1)acceleration of parallel computation due to modern powerful GPUs that make it possible to train models with large amount of parameters\cite{Chetlur2014CuDNN:Learning}, (2)larger datasets are released boosting more open source research and training, e.g., ImageNet\cite{Deng2009Imagenet:Database} , and (3) new efficient neural network structures, e.g., convolution neural network (CNN) which utilizes weight sharing and local connection\cite{LeCun1989BackpropagationRecognition}. In addition, deep learning methods are also successfully applied to the category of low-level vision problems including image denoising \cite{Xie2012ImageNetworks}, super resolution \cite{Dong2014LearningSuper-resolution}, and image restoration \cite{Mao2016ImageConnections}, etc., achieving state-of-the-art results. 

Although these methods mainly focus on natural image processing, several efforts have been made to apply these promising methods to medical image analysis. U-Net\cite{Ronneberger2015U-Net:Segmentation} is a fully convolutional network for medical image segmentation which consists of a contracting path and an expansive path to extract features at different resolution. To regain the lost resolution information, U-Net also employs skip connection to concatenate corresponding contracting and expansive steps. Inspired by U-Net, Han \textit{et al}\cite{Han2016DeepAnalysis} proposed a multi-scale CNN to remove streaking artifacts in sparse-view CT images, using residual learning. WaveNet\cite{Kang2016WaveNet:Reconstruction}, which is also used for low-dose X-ray CT reconstruction, adopts a similar structure combined with multi-scale wavelet transformation as feature augmentation for input data. In the field of low-dose PET reconstruction, compared with low-dose CT reconstruction, there are few researches on low-dose PET image denoising that utilize deep learning methods. Xiang \textit{et al}.\cite{XIANG2017} proposed a deep learning method to predict standard-dose PET images from low-dose PET images and corresponding MR T1 images with an auto-context convolution network which tries to refine the prediction results step by step. 

In terms of dose reduction factor (DRF), methods in \cite{Yu2016NoiseImages,DuttaJoyitaANDLeahy2013Non-LocalImages,Bagci2013DenoisingEstimate,Pogam2013DenoisingQuantitation} are used to denoising standard-dose images (DRF = 1) while methods in \cite{Wang2016PredictingRepresentation,Wang2017SemisupervisedMRI} try to reconstruct standard-dose images from quarter-dose images (DRF=4). However, to the best of our knowledge, there is no work that reconstructs low-dose PET images with higher DRF. 

In this paper, we propose a deep learning method to reconstruct standard-dose PET images from ultra-low-dose images (99.5\% reduction or DRF=200), using a fully convolutional encoder-decoder residual deep network model. To our best knowledge, this is the first time a deep learning method is proposed and demonstrated for enabling ultra-low-dose PET reconstruction at such a high reduction factor and with in-vivo PET datasets.

\section{Method}

\subsection{Dataset and experiments setup}

PET/MRI images from nine patients with glioblastoma (GBM) were acquired on a simultaneous time-of-flight enabled PET/MRI system (SIGNA, GE Healthcare) with standard dose of 18F-fluorodeoxyglucose (FDG) (370 MBq). Images were acquired for about 40 min, beginning 45 min after injection.  We stored the raw count listmode datasets for each scan and then generate synthesized low-dose raw data at $\rm{DRF}=200$ by simply randomly selecting $0.5\%$ of the count events, spread uniformly over the entire acquisition period. Then we reconstruct PET images from the acquired data at $\rm{DRF}=1$ (standard full dose) and $\rm{DRF}=200$ (target low dose) using standard OSEM methods (28 subsets, 2 iterations).

The size of each reconstructed 3D PET data is $256\times256\times89$. There are slices of air at the top and bottom, which are removed. To avoid over fitting, data augmentation is adopted during the training process to simulate a larger dataset. Before fed into the network, images are randomly flipped along x and y axes and transposed. 

\subsection{Deep Learning based low-dose PET reconstruction}
The goal of this work is to train a model to learn to reconstruct from the $\rm{DRF}=200$ image to $\rm{DRF}=1$ reconstruction.

\subsubsection{Network structure}

As shown in Fig. \ref{fig_network_structure}, the proposed fully convolutional network is based on an encoder-decoder structure with symmetry concatenate connections between corresponding stages, which is inspired by the U-Net structure but modified for image synthesis task instead of segmentation task. Specifically, each stage consists of convolution with 3$\times$3 kernels, batch normalization, and rectified linear unit (ReLU). The downsampling and upsampling between stages are done by 2$\times$2 max pooling and bilinear interpolation respectively. By downsampling and then upsampling the image, the network can extract the multi-scale and high-level features from the image. The low-dose PET image reconstruction task is similar to image denoising which is within the category of low-level vision problems and are susceptible to resolution loss if only a encoding-decoding procedure is used. Therefore, concatenate connections are added to preserve local information and resolution of the image.

\subsubsection{Residual learning}

Residual learning\cite{He2016DeepRecognition} is first introduced into CNN as a technique to avoid performance degradation when training very deep CNNs. 
It shows by separating the identity and the residual part, the neural network can be trained more effectively and efficiently. Originally, residual learning is used in image recognition task \cite{He2016DeepRecognition} and Zhang \textit{et al.} \cite{Zhang2017BeyondDenoising} proposed DnCNN which is the first denoising convolution network using residual learning. It is showed in \cite{Han2016DeepAnalysis}, using persistent homology analysis, that the residual manifold of CT artifacts has a much simpler structure. Our network also employs the residual learning technique, by adding a residual connection from input to output directly, i.e., instead of learning to generate standard-dose PET images directly, the network tries to learn the difference between standard-dose images outputs and low-dose images inputs. Our study shows that residual learning can also lead to a significant improvement in network performance for low-dose PET reconstruction problem.

\subsubsection{Using multi-slice as input}
\begin{figure}[thpb]
\centerline{\includegraphics[width=0.5\textwidth]{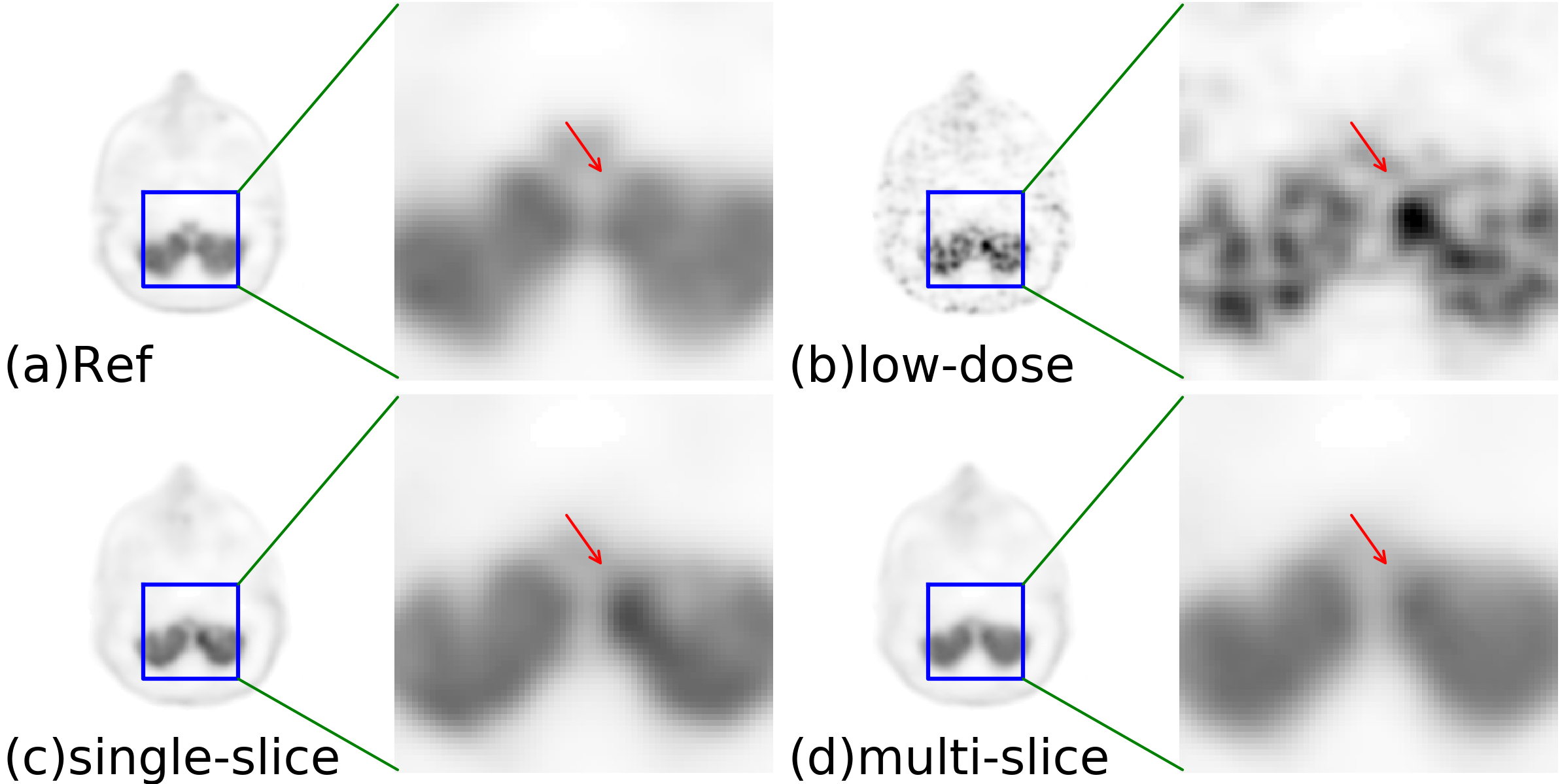}}
\caption{Effect of multi-slice input. (a) standard-dose, (b) 200x low-dose, (c) reconstructed result using single-slice input, and (d) reconstructed result using multi-slice input (3 slices).}\label{fig_eg_multislice}
\label{fig}
\end{figure}

Using only the low-dose image as input for the neural network may not provide enough information to reconstruct the standard-dose counterpart. As shown in fig, the noise due to dose reduction cannot be fully eliminated by the network for the network may have insufficient information to distinguish noise from brain structure. To address this problem, we use multi-slice input instead of single-slice input, i.e., adjoining slices are stacked as different input channels. In general, the multi-slice inputs can be regarded as a kind of feature augmentation. Since the structure of the brain is deterministic, adjoining slices may share similar structure while having different noise which is random. Thus, combining different slices as input can provide the network with 2.5D structural information that can be used to distinguish random noise from the consistent structure. One example is illustrated in Fig. \ref{fig_eg_multislice}, in the low-dose PET image there is a black noise in the zoomed part, which cannot be eliminated but hallucinated as a structure by the network trained with single-slice input. However, the network trained with three-slice input can achieve better results, as shown in Fig. \ref{fig_eg_multislice}(d). Training 2.5D multi-slice inputs is different from training with 3D convolution network since the former solution sues depth-wise operation of 3D convolution which has fewer parameters and higher efficiency.

\subsubsection{Selection of loss functions}
The mean squared error (MSE) or $L_2$ loss is still the most popular choice of loss function in training networks for image restoration problems, e.g., super resolution or denoising. The use of MSE as loss function is under the assumption of additive white Gaussian noise, which should be independent of the local features of the image. However, this is not valid for low-dose PET reconstruction in general. Since the intensity of PET image reflects the activity distribution of tracer in the subject and the noise results from dose reduction is related to the counting of each detector, noise and spatial information are not independent. Besides, the MSE loss may be not suitable for task related to clinical evaluation for it relates poorly to the human visual system and produces splotchy artifacts\cite{Zhao2017LossNetworks}. 

Besides the traditional MSE, there are other loss functions  that can be used to measure image similarity between reconstructed image and the ground-truth image. The $L_1$ loss is the mean absolute error of two images which can be defined as

\begin{equation}
L^{l_1}=\frac{1}{NM}\sum_{i=1}^N\sum_{j=1}^M|x_{ij} - y_{ij}|
\end{equation}
where $N$, $M$ are number of rows and columns of the image respectively, while $x_{ij}$ and $y_{ij}$ denote the intensity at pixel $(i,j)$ in the two images. To measure the structural and perceptual similarity, structural similarity index (SSIM\cite{Wang2004ImageSimilarity}), and multi-scale the structural similarity index (MS-SSIM\cite{Wang2003MultiscaleAssessment}) are proposed and can be estimated as 

\begin{equation}
L^{\rm{SSIM}} = \frac{1}{NM}\sum_{i=1}^N\sum_{j=1}^M (1 - \mbox{SSIM}(i,j))
\label{eq:SSIM}
\end{equation}

\begin{equation}
L^{\mbox{MS-SSIM}} = \frac{1}{NM}\sum_{i=1}^N\sum_{j=1}^M( 1 - \mbox{MS-SSIM}(i,j))
\end{equation}

where
\begin{equation}\label{eq_ssim}
\begin{aligned}
\mbox{SSIM}(i,j) = &\frac{(2\mu_x\mu_y+C_1)}{(\mu_x^2+\mu_y^2+C_1)}*\frac{(2\sigma_{xy}+C_2)}{(\sigma_x^2+\sigma_y^2+C_2)}\\
= &l(i,j)*cs(i,j)
\end{aligned}
\end{equation}

\begin{equation}
\mbox{MS-SSIM}(i,j) = l_{K}(i,j) * \prod_{k=1}^K cs_j(i,j)
\end{equation}
$C_1$ and $C_2$ are constants. $\mu_x$, $\mu_y$, $\sigma_x$, $\sigma_y$, and $\sigma_{xy}$ are the image statistics calculated in the patch centered at pixel $(i,j)$. $K$ is the number of level of multi-scale.

recent researches\cite{Zhao2017LossNetworks,Ridgeway2015LearningMetrics} suggested that $L_1$, SSIM, MS-SSIM are more perceptually preferable in image generative model. Among these three alternatives, the L1 loss can not only avoid the patchy artifact brought by L2 loss but add almost no overhead in back propagation compared with SSIM and MS-SSIM. Therefore, the L1 loss is selected as a loss function for training procedure in the following experiments.

\subsection{Computation environment and hardware settings}
All the computation works were done on a Ubuntu server with 2 NVIDIA GTX 1080Ti GPUs. The proposed network is implemented in TensorFlow. The RMSprop optimizer is used in our experiments with a learning rate initialized by $1\times10^{-3}$ which slowly decreases down to $2.5\times10^{-4}$. The network was trained for 120 epochs. Convolution kernel were initialized with truncated Gaussian distributions with zero mean and standard deviation 0.02. All biases are initialized with zero.

\subsection{Evaluation and similarity metrics}

To evaluate the performance of the proposed method and demonstrate it can generalize for new dataset, especially for new patient data with different pathology, we used the leave-one-out cross validation (LOOCV). For each of the patient dataset, we generated the full-dose reconstruction using the model trained only on the other eight patients. We used the statistics of LOOCV results to quantify the generalization error of the proposed model.
To quantitatively evaluate image quality, three similarity metrics are used in our experiment, including the normalized root mean square error (NRMSE), peak signal to noise ratio (PSNR) and SSIM. SSIM is defined in equation \ref{eq_ssim}, while NRMSE and PSNR are defined as follows.

\begin{equation}
\mbox{NRMSE} 
= \sqrt{\frac{\sum_{i=1}^N\sum_{j=1}^M(x_{ij} - y_{ij})^2}{\sum_{i=1}^N\sum_{j=1}^M y_{ij}^2}}
\end{equation}

\begin{equation}
\rm{PSNR} = 20*\mbox{log}_{10}(\frac{\rm{MAX}}{\sqrt{\rm{MSE}}})
\end{equation}

where MAX is the is the peak intensity of the image.
To better match the metric computation to the real clinical assessment, all the similarity metrics were computed after applying a brain mask estimated using image support. 

\section{Results}

\subsection{Comparison with other methods}
We compared our proposed method against three state-of-the-art denoising methods in low-dose PET reconstruction, including NLM \cite{Buades2005ADenoising}, BM3D \cite{Dabov2006ImageFiltering} and auto-context network \cite{XIANG2017} (AC-Net). Cross validation is conducted to evaluate these methods. 

\begin{figure*}[thpb]
\centering
\includegraphics[width = 0.9\textwidth]{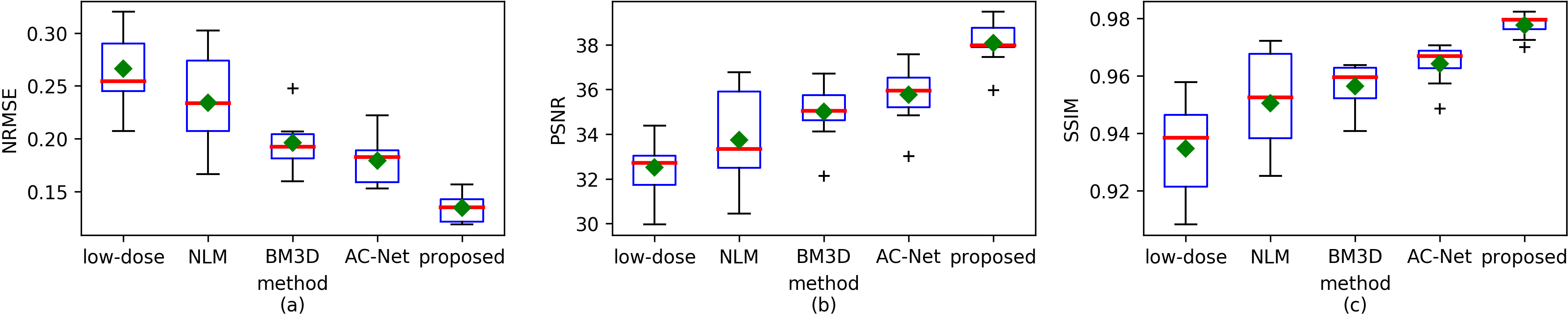}
\caption{Comparison of the averaged performance and similarity metrics of different methods for low-dose reconstruction, where green diamonds denote means.}\label{method}
\end{figure*}

\begin{figure*}[thpb]
\centering
\includegraphics[width = 0.9\textwidth]{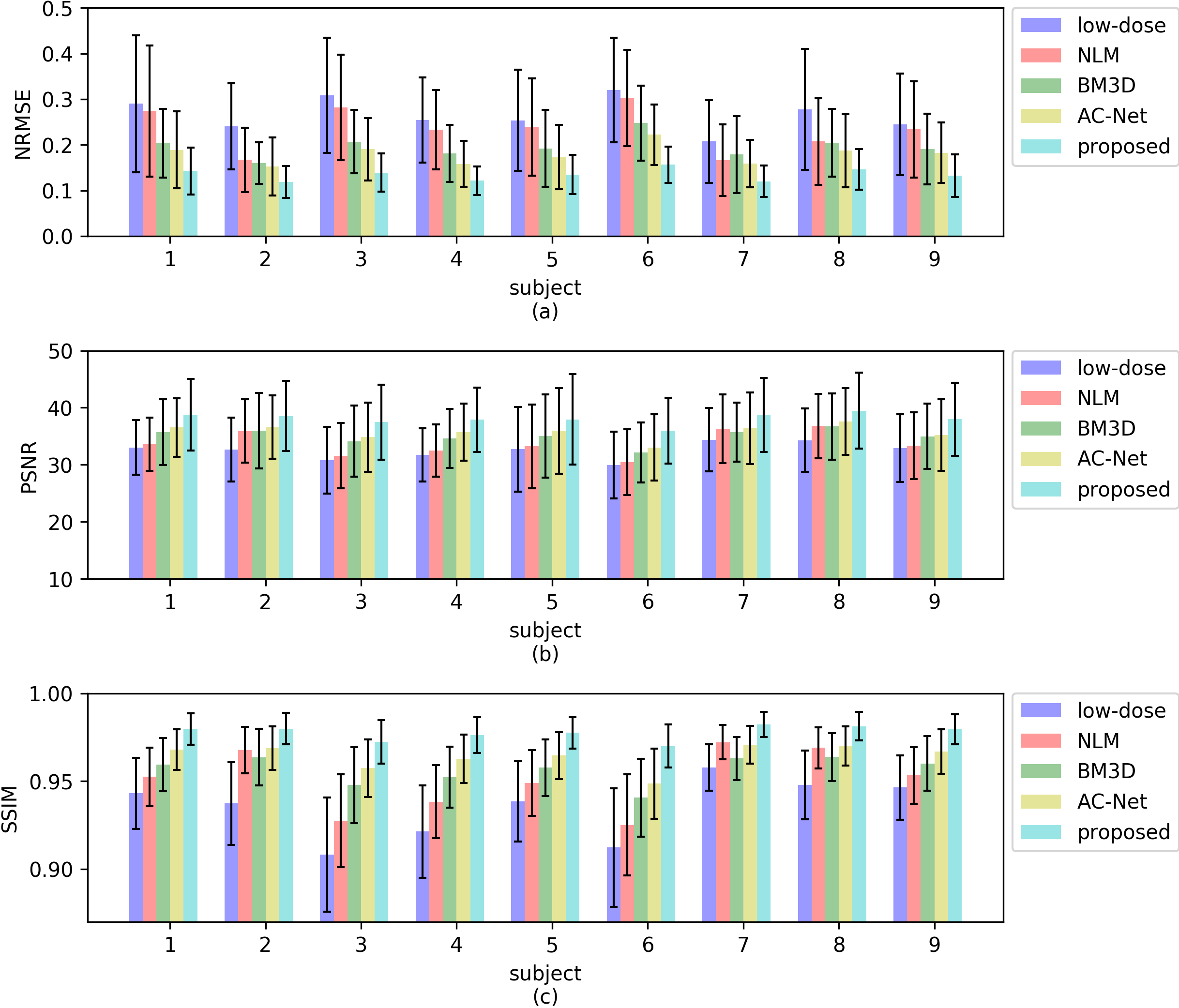}
\caption{Quantitative comparison between our proposed method and previous method using LOOCV.}\label{method_bar}
\end{figure*}

Fig. \ref{method} shows the average performance in terms of NRMSE, PSNR and SSIM of all the subjects, while Fig. \ref{method_bar} gives the scores of these three metrics for all 9 subjects in the leave-one-out testing. 

\begin{figure*}[thpb]
\centering
\includegraphics[width = \textwidth]{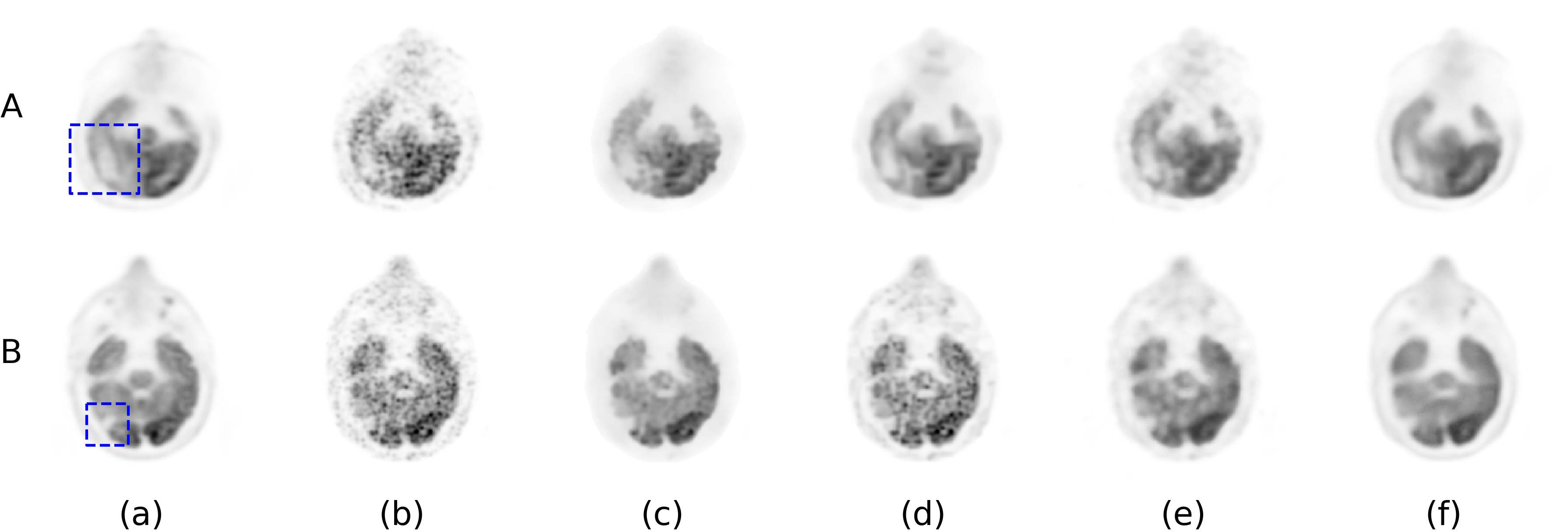}
\caption{Results from different methods for comparison. (a) standard-dose, (b) low-dose, (c) NLM, (d)BM3D, (e) AC-Net, and (f) proposed.}\label{method_vis}
\end{figure*}

\begin{figure*}[thpb]
\centering
\includegraphics[width = \textwidth]{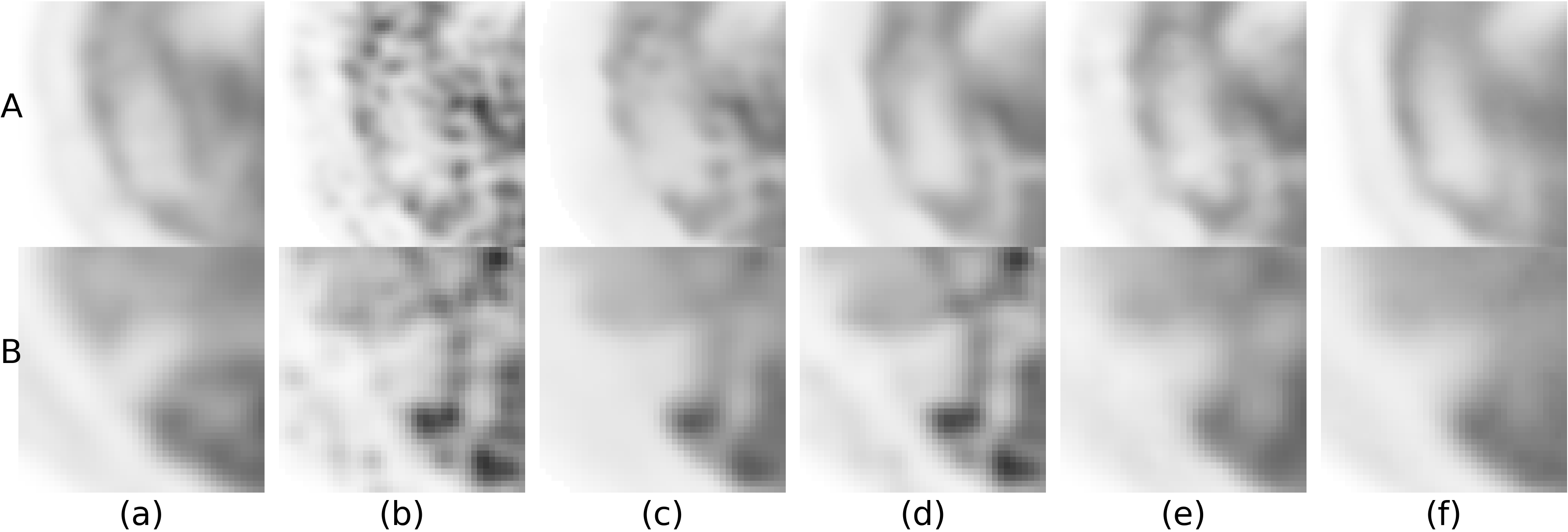}
\caption{Zoomed part in Fig. \ref{method_vis}. (a) standard-dose, (b) low-dose, (c) NLM, (d)BM3D, (e) AC-Net, and (f) proposed.}\label{method_vis_zoom}
\end{figure*}

\begin{figure*}[thpb]
\centering
\includegraphics[width = \textwidth]{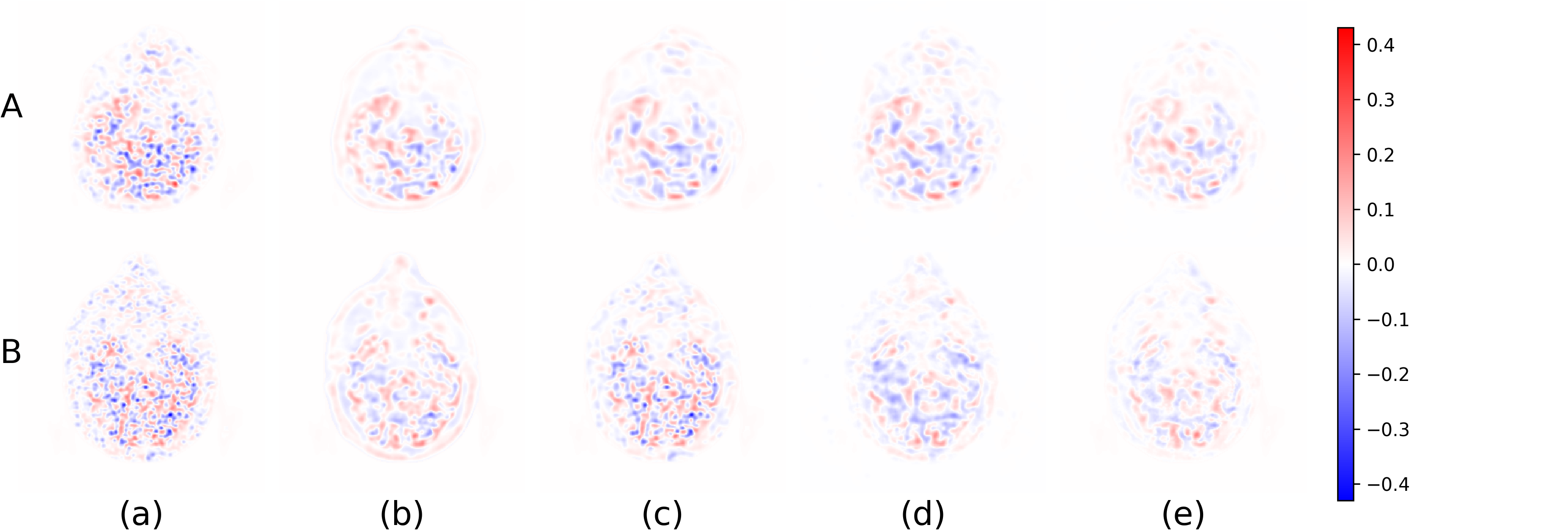}
\caption{Error map of Fig. \ref{method_vis}. (a) low-dose, (b) NLM, (c)BM3D, (d) AC-Net, and (e) proposed.}\label{method_vis_err}
\end{figure*}

To examine perceptual image quality, two representative slices are selected from different subjects. The quantitative metrics in terms of NRMSE, PSNR and SSIM of the selected slices are listed in Table \ref{table_vis_metric}. The reconstruction results, zoomed tumors, and the corresponding error map are visually illustrated in Fig. \ref{method_vis}, Fig. \ref{method_vis_zoom}, and Fig. \ref{method_vis_err} respectively. 

\begin{table}[htpb]
\caption{QUANTITATIVE RESULTS ASSOCIATED WITH DIFFERENT ALGORITHMS FOR REPRESENTAIVE SLICES.}
\label{table_vis_metric}
\begin{center}
\begin{tabular}{c c c c c c c}
\toprule
 & & slice A & & & slice B & \\

 & NRMSE & PSNR & SSIM & NRMSE & PSNR & SSIM\\
\midrule
low-dose & 0.228 & 30.05 & 0.917 & 0.214 & 29.58 & 0.899\\
NLM      & 0.153 & 33.50 & 0.958 & 0.142 & 33.13 & 0.947\\
BM3D     & 0.145 & 32.98 & 0.961 & 0.189 & 30.70 & 0.926\\
AC-Net   & 0.147 & 33.83 & 0.960 & 0.137 & 33.50 & 0.958\\
Proposed & \textbf{0.124} & \textbf{35.33} & \textbf{0.974} & \textbf{0.106} & \textbf{35.66} & \textbf{0.969}\\
\bottomrule
\end{tabular}
\end{center}
\end{table}

\subsection{Contributions of skip connections}
\begin{figure}[thpb]
\centering
\includegraphics[width=0.45\textwidth]{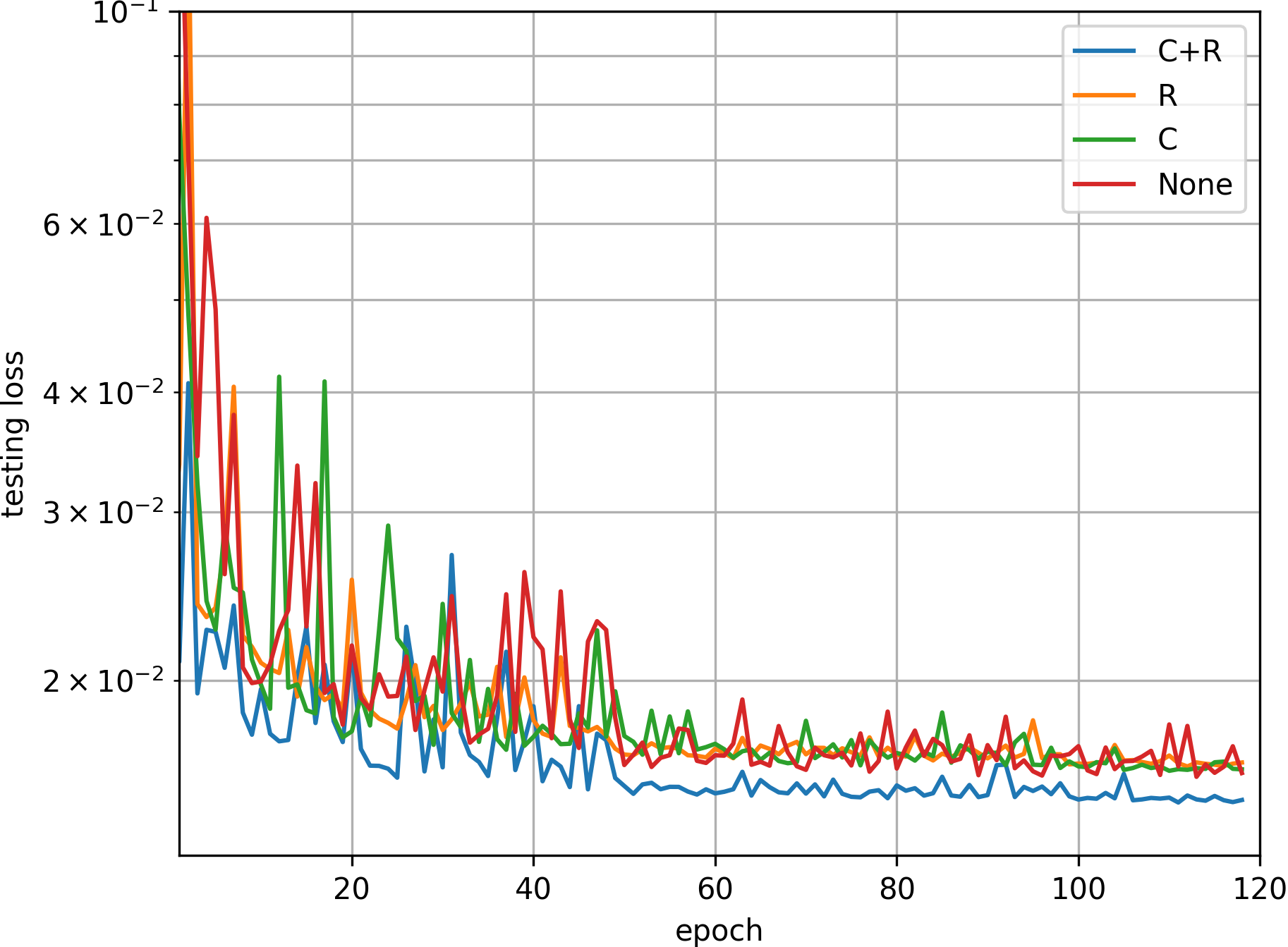}
\caption{Testing loss of networks with different types of skip connection. R means residual connection and C means concatenate connection.}\label{skip_val_loss}
\end{figure}

\begin{figure*}[thpb]
\centering
\includegraphics[width=0.9\textwidth]{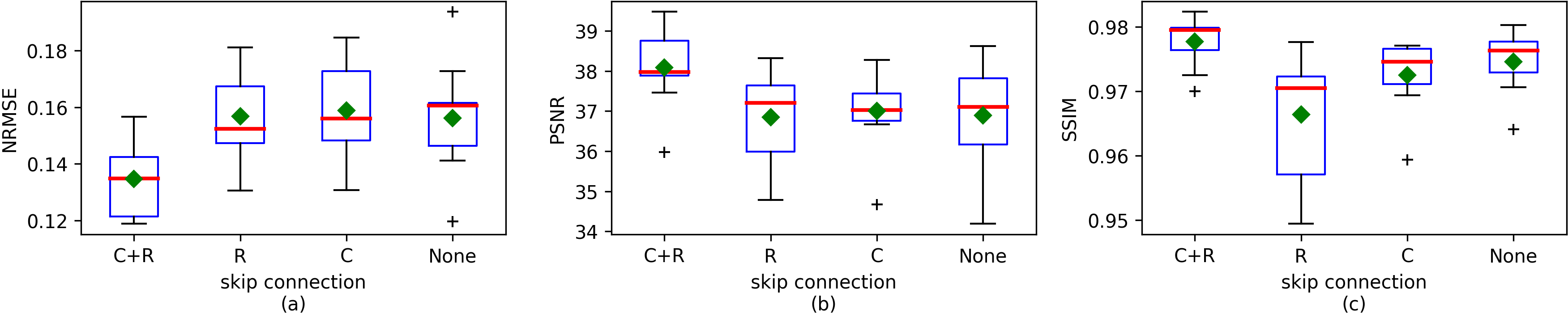}
\caption{Similarity metrics of network with different types of skip connection. R means residual connection and C means concatenate connection.}\label{skip}
\end{figure*}

We explored the contribution to reconstruction from different skip connection components in the network. There are two types of skip connections in our proposed network. One is the residual connection from input to output, and the other is the concatenating connections between corresponding encoder and decoder layers. To evaluate the effect of these two types of skip connection on the network performance, four different models are trained and tested, i.e., (1) with both types of skip connection, (2) with only concatenate connection, (3) with only residual connection, and (4) without any skip connection. Fig. \ref{skip_val_loss} shows the different testing loss of these four models during training and the quantitative results of cross validation are illustrated in Fig. \ref{skip}. 

\subsection{Contributions of multi-slice inputs}

\begin{figure}[thpb]
\centering
\includegraphics[width=0.4\textwidth]{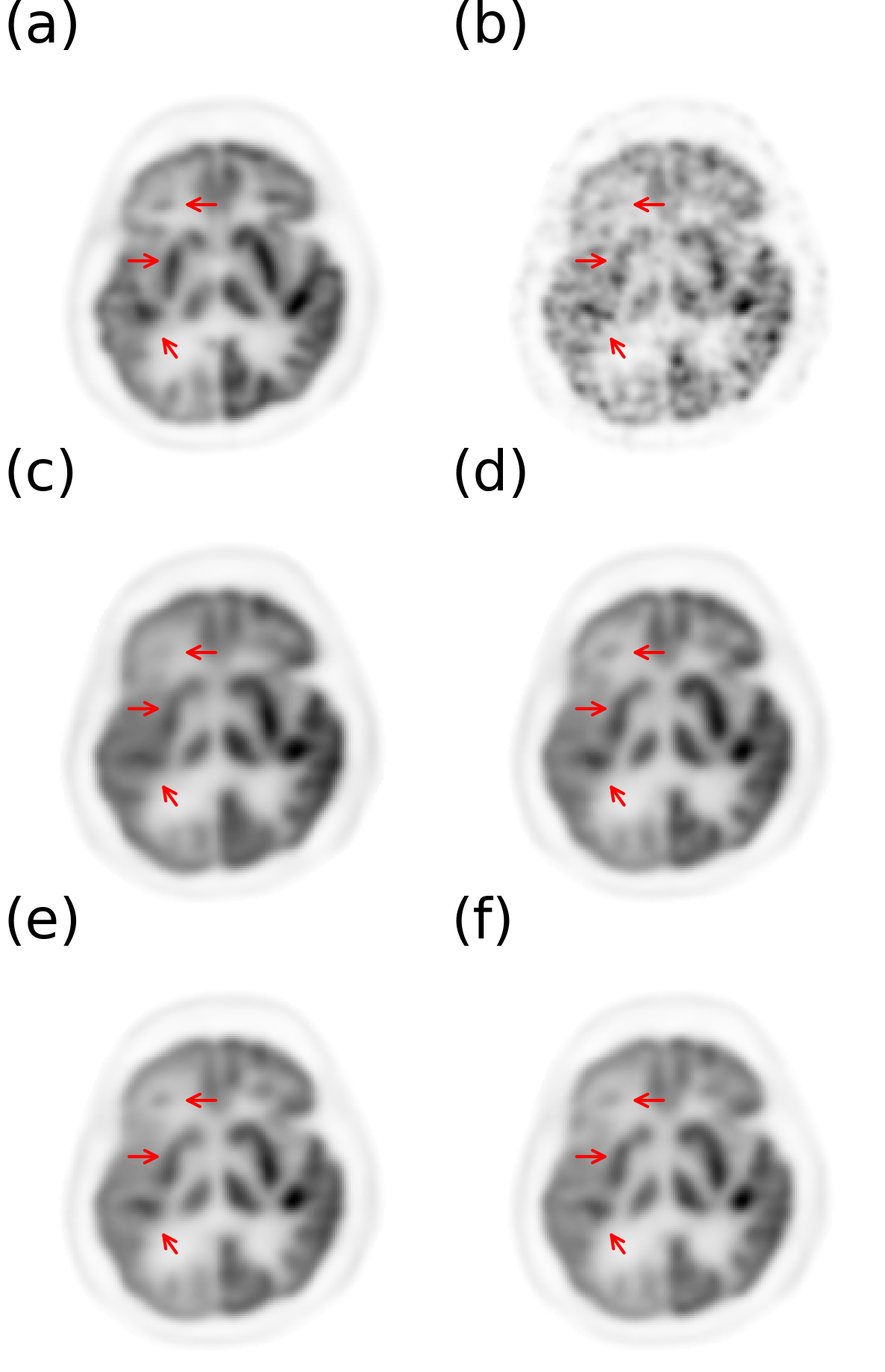}
\caption{Reference, inputs and reconstruction results using model with different settings in multi-slice inputs. (a) standard-dose, (b) low-dose, (c) single slice, (d) three slices, (e) five slices, and (f) seven slices}\label{multislice__vis}
\end{figure}

As mentioned above, we used multi-slice input to combine information from adjoining slices so that the network can more accurately generate reconstruction with less noise and artifact while robustly preserve original structure and details. To study the limit of this technique, networks with different numbers of input slices (1,3,5,7) are trained and their results are compared, shown in Fig. \ref{multislice__vis}. 

\begin{figure*}[thpb]
\centering
\includegraphics[width=0.9\textwidth]{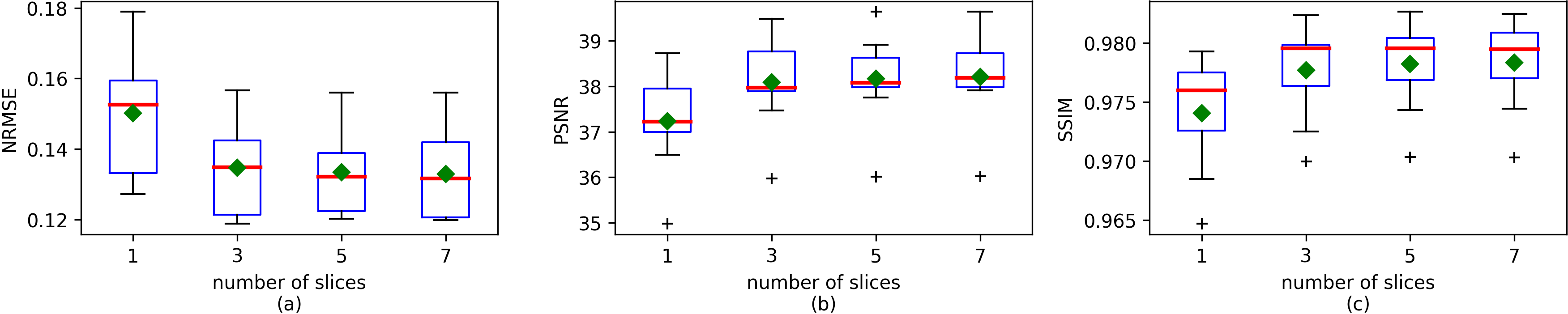}
\caption{Similarity metrics for networks trained with different numbers of input slices}\label{multislice}
\end{figure*}

Fig. \ref{multislice} shows results of three similarity metrics of networks trained with different numbers of slices as 2.5D inputs in the network. The evolutions of the three metrics all validate the performance gain of the proposed method using more input slice number. Compared with single-slice input, three-slice input can provide significantly better results. However, the performance gain of the network, by continuously adding slices more than 3 slices, is not as significant. Similar phenomenon can be seen in Fig\ref{multislice__vis}. (d)-(f) contain details that are missing or blurred in (c). However, (d)-(f) are perceptually similar.

\subsection{Depth of network}

\begin{figure*}[thpb]
\centering
\includegraphics[width=0.9\textwidth]{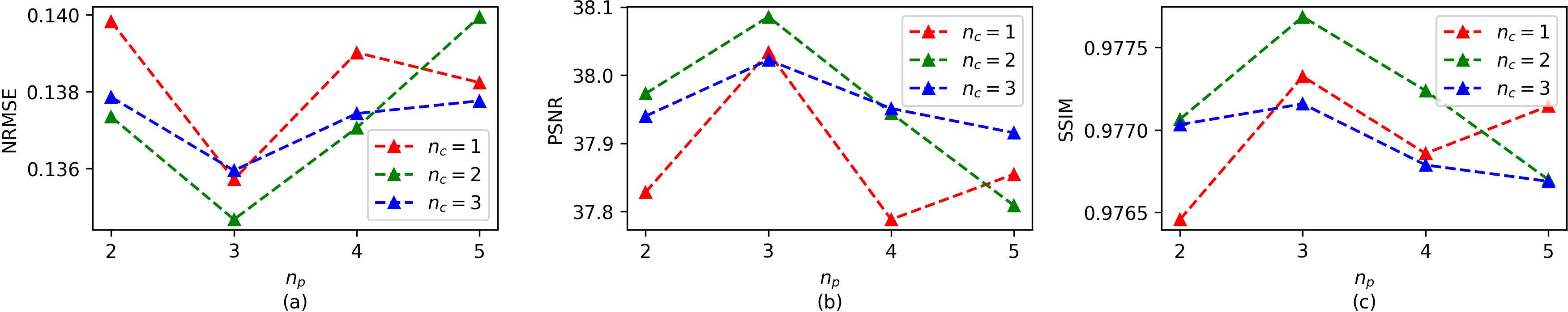}
\caption{Performance of networks without different depth evaluated with average NRMSE, PSNR and SSIM over all the 9 subjects in our study.}\label{depth}
\end{figure*}

To optimize our proposed network, experiments are conducted to evaluate the impact of depth of our model on network performance. Two hyper-parameters are used to control the depth of our network, namely number of pooling layers ($n_p$) and number of convolutions between two poolings ($n_c$). The strategy of grid search is adopted. In our experiment, $n_p$ varies from 2 to 5 while $n_c$ varies from 1 to 3. The results are shown in Fig. \ref{depth}, which suggest that $n_p=3$ and $n_c=2$ is the best architecture in our study.

\section{Discussion}

\subsection{Compare with other methods}
Quantitative results in Fig. \ref{method} and Fig. \ref{method_bar} show that the propose method demonstrated the best performance in all nine subjects in the data set, compared with other methods that we have tested. From the visual results, it also suggests that our proposed method have highest image quality. NLM produces patchy artifact in the image as shown in Fig \ref{method_vis}(c). Both BM3D and AC-Net cannot fully remove the noise in low-dose image and tend to over-blur the image without recover important details, as illustrated in Fig. \ref{method_vis}(d) and (e). Same conclusion can also be drawn from the error map in Fig. \ref{method_vis_err}. In addition, our proposed method can achieve the best perceptual result in the region of GBM, as shown in Fig. \ref{method_vis_zoom}.

In terms of computational costs, although deep learning requires long time for training, their efficiency in inference can easily outperforms traditional methods due to efficient implementation with Tensorflow and parallelization on GPUs. Time consumptions of each method for a $256\times 256$ image are listed in Table \ref{table_test_time}. Compared with other methods, the proposed solution is not only more accurate but also more efficient.

It is the encoder-decoder structure that enable the network to adopt more parameters and channels to extract higher level features while reducing computation time, compared with single-scale model used in AC-Net.

\begin{table}[htpb]
\caption{TESTING TIME (Per Image) FOR EACH METHOD.}
\label{table_test_time}
\begin{center}
\begin{tabular}{c c c}
\toprule
Method & Average Speed/Image (ms) \\
\midrule
NLM(CPU) & 1180 \\
NLM(GPU) & 63 \\
BM3D(CPU) & 680 \\
BM3D(GPU) & 232 \\
AC-Net(GPU) & 27 \\
Proposed(GPU) & \textbf{19}\\
\bottomrule
\end{tabular}
\end{center}
\end{table}

\subsection{Benefits from concatenation and residual skip connections}

As the result shown in Fig. \ref{skip}, the model with both types of skip connections obviously achieves the best performance. However, for model with only one type of skip connection, their performances are close to that of the model without skip connection, or even worse. There results indicates that these two kinds of connections are not independent.

\subsection{Benefits from 2.5D augmentation}
Here we compared both quantitatively and qualitatively on the reconstruction using different options for combining multi-slice inputs. Detailed structures in Fig.\ref{multislice__vis}(c) are blurred during the denoising process, while they are preserved in Fig. \ref{multislice__vis}(d)-(f), which shows the benefits from multi-slice inputs.

Since resolution of our 3D PET data along z axial direction is worse than within axial image. Stacking a few slices along z axis can recover the 3D spatial relationship. 
Here we showed significant performance improvement from the 2.5D slice with augmentation by only using 3 slices, however the performance is not further improved by using more slices as inputs. This result is consistent with the assumption that the structural similarity of different slices persists until the relationship and redundancy one can leverage between slices vanish eventually due to distance. 

\section{Conclusion}

In this paper, a deep fully convolutional network was proposed for ultra-low-dose PET reconstruction, where multi-scale encoder-decoder architecture, concatenate connections and residual learning are adopted. 

The results showed the proposed method has superior performance in reconstructing high-quality PET images and generating comparable quality as from normal-dose PET images. The method significantly reduces noise while robustly preserve resolution and detailed structures. 

In addition, we demonstrated how different components of the proposed method contribute to the improved performance: the design of loss function, 2.5D multi-slice inputs as well as concatenating and residual skip connections, etc. Detailed quantitative and qualitative comparison proved the proposed method can better preserve structure and avoid hallucination due to noise and artifacts. 

With extensive comparison, our proposed method achieves significantly better reconstruction compared with previous methods from ultra-low-dose PET data from $0.5\%$ of the regular dose, potentially enabling safer and more efficient PET scans.

\section{Acknowledge}
The authors would like to thank Dr. Mohammad Mehdi Khalighi for his technical assistance in data acquisition and processing.

\bibliographystyle{ieeetr}
\bibliography{Mendeley.bib}



\addtolength{\textheight}{-12cm}   


\end{document}